\documentclass{article}
\usepackage{ijcai18}

\usepackage{times}
\usepackage{xcolor}
\usepackage{soul}
\usepackage[utf8]{inputenc}
\usepackage[small]{caption}

\usepackage{graphicx}
\usepackage{amsmath}
\usepackage{multirow}
\usepackage{algorithm}
\usepackage{algorithmic}
\newcommand{\ignore}[1]{{}}

\newcommand{\sgn}{\mathop{\mathrm{sgn}}}

\def\ours{{\tt EASDRL}}
\def\w{\mathbf{w}}
\def\X{\mathbf{X}}

\title{Extracting Action Sequences from Texts Based on Deep Reinforcement Learning}

\author{
Wenfeng Feng$^1$, 
Hankz Hankui Zhuo$^1$\thanks{Corrsponding Author}, 
Subbarao Kambhampati$^2$ 
\\ 
$^1$ School of Data and Computer Science, Sun Yat-sen University, Guangzhou, China \\
$^2$ Department of Computer Science and Engineering, Arizona State University, Tempe, Arizona, US \\
fengwf@mail2.sysu.edu.cn, 
zhuohank@mail.sysu.edu.cn, 
rao@asu.edu
}

\begin{document}
\maketitle
\begin{abstract}
Extracting action sequences from  natural language texts is
challenging, as it  requires commonsense inferences based on world
knowledge. Although there has been work on extracting action scripts,
instructions, navigation actions, etc., they require that either the
set of candidate actions be provided in advance, or that action
descriptions are restricted to a specific form, e.g., description
templates. In this paper, we aim to extract action sequences from
texts in \emph{free} natural language, i.e., without any restricted
templates, provided the candidate set of actions is unknown. We
propose to extract action sequences from texts based on the deep
reinforcement learning framework. Specifically, we view ``selecting''
or ``eliminating'' words from texts as ``actions'', and the texts associated with actions as ``states''. We then build Q-networks to learn the policy of extracting actions and extract plans from the labeled texts. We demonstrate the effectiveness of our approach on several datasets with comparison to state-of-the-art approaches, including online experiments interacting with humans.
\end{abstract}

\section{Introduction}

AI agents
will increasingly find  assistive roles in
homes, labs, factories  and public places. The widespread adoption of
conversational agents such as Alexa, Siri and Google Home demonstrate
the natural demand for such assistive agents. 
To go beyond supporting the  simplistic ``what is the weather?''
queries however, these agents need domain-specific
knowledge such as the recipes and standard operating procedures. While
it is possible to hand-code such knowledge (as is done by most of the
``skills'' used by Alexa-like agents), ultimately that is too labor
intensive an option. One idea is to have these agents automatically ``read''
instructional texts, typically written for human workers, and
convert them into action sequences and plans for later use (such as learning domain models \cite{DBLP:journals/ai/ZhuoM014,DBLP:journals/ai/Zhuo014} or model-lite planning \cite{DBLP:journals/ai/ZhuoK17}). 
Extracting action sequences from natural language texts meant for
human consumption is however challenging, as it  requires agents to understand
complex contexts of actions.

\begin{figure}[!ht]
\begin{center}
\includegraphics[width=0.49\textwidth]{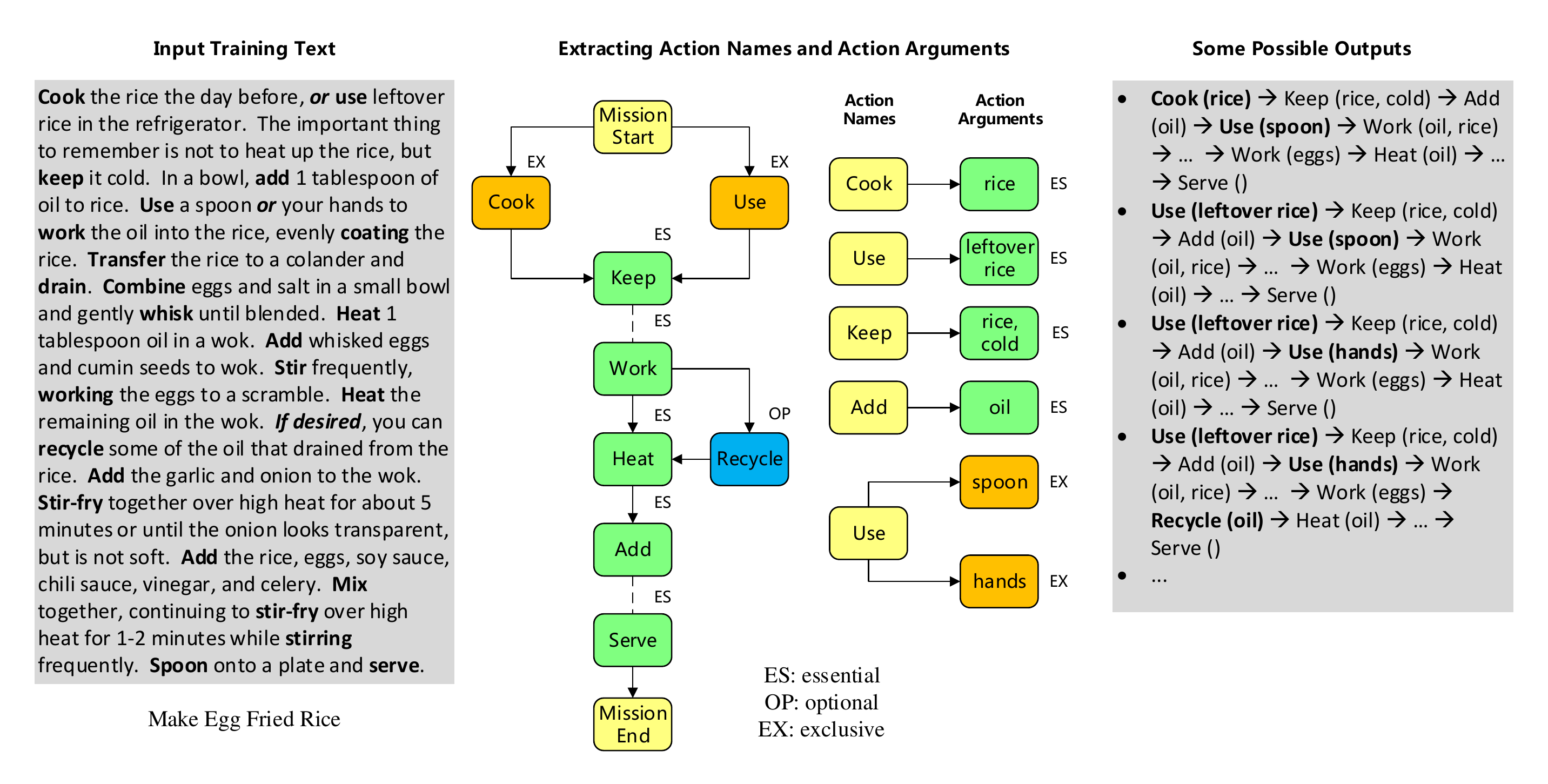}
\caption{Illustration of our action sequence extraction problem}
\label{examples}
\end{center}
\end{figure}

For example, in Figure \ref{examples}, given a document of action
descriptions (the left part of Figure \ref{examples}) such as
``\emph{Cook the rice the day before, or use leftover rice in the
  refrigerator. The important thing to remember is not to heat up the
  rice, but keep it cold.}'', which addresses the procedure of making
egg fired rice, an action sequence of ``\emph{cook(rice), keep(rice,
  cold)}'' or ``\emph{use(leftover rice), keep(rice, cold)}'' is
expected to be extracted\ignore{ from the action descriptions}. This
task is challenging. For \ignore{example, for }the first sentence, the
agent needs  to learn to figure out that ``cook'' and ``use'' are
\emph{exclusive} (denoted by ``EX'' in the middle of Figure
\ref{examples}), meaning that we could extract only one of them; for
the second sentence, we need to learn to understand that among the
three verbs ``remember'', ``heat'' and ``keep'', the last one is the
best because the goal of this step is to ``keep the rice cold''
(denoted by ``ES'' indicating this action is \emph{essential}). There
is also another action ``Recycle'' denoted by ``OP'' indicating this
action can be extracted \emph{optionally}. We also need to consider
action arguments which can be either ``EX'' or ``ES'' as well (as
shown in the middle of Figure \ref{examples}). The possible action
sequences extracted are shown in the right part of Figure
\ref{examples}. This action sequence extraction problem is different
from sequence labeling and dependency parsing, since we aim
to extract ``meaningful'' or ``correct'' action sequences (which
suggest some actions should be ignored because they are exclusive),
such as ``\emph{cook(rice), keep(rice, cold)}'', instead of
``\emph{cook(rice),use(leftover rice), remember(thing), heat(rice),
  keep(rice, cold)}'' as would be extracted by LSTM-CRF models\cite{DBLP:conf/acl/MaH16} or external NLP tools.

There has been work on extracting action sequences from action
descriptions. For example, \cite{DBLP:conf/acl/BranavanCZB09} propose
to map instructions to sequences of executable actions using
reinforcement learning. \cite{Mei2016Listen,Daniele2017Navigational}
interpret natural instructions as action sequences or generate
navigational action description using an encoder-aligner-decoder
structure. Despite the success of those approaches, they all require a
limited set of action names given as input, which are mapped to action
descriptions. Another approach, proposed by
\cite{DBLP:conf/aips/LindsayRFHPG17}, builds action sequences from
texts based on dependency parsers and then builds planning models,
assuming texts are in restricted templates when describing actions.


In this paper, we aim to extract meaningful action sequences from
texts in \emph{free} natural language, i.e., without any restricted
templates, even when the candidate set of actions is unknown. We
propose an approach called {\ours}, which stands for
\textbf{E}xtracting \textbf{A}ction \textbf{S}equences from texts
based on \textbf{D}eep \textbf{R}einforcement \textbf{L}earning.  In
 {\ours}, we view texts associated with actions
as ``states'', and associating words in texts with labels as
``actions'', and then build deep Q-networks to extract action
sequences from texts. We capture complex relations among actions
by considering previously extracted actions as parts of states for
deciding the choice of next operations. In other words, once we know
action ``cook(rice)'' has been extracted and included as parts of
states, we will choose to extract next action ``keep(rice, cold)''
instead of ``use(leftover rice)'' in the above-mentioned example.

In the remainder of paper, we first review previous work related to our approach. After that we give a formal definition of our plan extraction problem and present our {\ours} approach in detail. We then evaluate our {\ours} approach with comparison to state-of-the-art approaches and conclude the paper with future work.

\section{Related Work}
There have been approaches related to our work besides the ones we mentioned in the introduction section. Mapping SAIL route instructions \cite{Macmahon2006Walk} to action sequences has aroused great interest of in natural language processing community. Early approaches, like \cite{DBLP:conf/aaai/ChenM11,DBLP:conf/acl/Chen12,Kim2013Unsupervised,Kim2013Adapting}, largely depend on specialized resources, i.e. semantic parsers, learned lexicons and re-rankers. Recently, LSTM encoder-decoder structure \cite{Mei2016Listen} has been applied to this problem and gets decent performance in processing single-sentence instructions, however, it could not handle multi-sentence texts well. 

There is also a lot of work on learning STRIPS representation actions \cite{DBLP:conf/ijcai/FikesN71,DBLP:conf/ki/PomarlanKB17} from texts. \cite{DBLP:conf/aaaifs/SilHY10,DBLP:conf/ranlp/SilY11} learn sentence patterns and lexicons or use off-the-shelf toolkits, i.e., OpenNLP\footnote{https://opennlp.apache.org/} and Stanford CoreNLP\footnote{http://stanfordnlp.github.io/CoreNLP/}. \cite{DBLP:conf/aips/LindsayRFHPG17} also build action models with the help of LOCM \cite{cresswell2009acquisition} after extracting action sequences by using NLP tools. These tools are trained for universal natural language processing tasks, they cannot solve the complex action sequence extraction problem well, and their performance will be greatly affected by POS-tagging and dependency parsing results. In this paper we aim to build a model that learns to directly extract action sequences without external tools. 

\ignore{
Recently \cite{Mnih2015Human} applied deep neural network (DQN) to solve
reinforcement learning problems and obtain state-of-the-art
results. \cite{Silver2016Mastering,kulkarni2016hierarchical} propose
DQN models to play more challenging games with enormous search space
or sparse
feedback. \cite{Lillicrap2015Continuous,Schaul2015Prioritized} develop
DQN structure to continuous action space and improve the experience
replay trick. Unlike those works which mainly focus on video games, \cite{DBLP:conf/emnlp/NarasimhanKB15,DBLP:conf/acl/HeCHGLDO16} take as input texts descriptions of games, which also shed  light on our task.}

\section{Problem Definition}\label{def}
Our training data can be defined by $\Phi=\{\langle X,Y\rangle\}$, where $X=\langle w_1,w_2,\ldots,w_N\rangle$ is a sequence of words and $Y=\langle y_1,y_2,\ldots,y_N\rangle$ is a sequence of annotations. If $w_i$ is not an action name, $y_i$ is $\emptyset$. Otherwise, $y_i$ is a tuple $(ActType, \{ExActId\}, \{\langle ArgId, ExArgId\rangle\})$ to describe \emph{type} of the action name and its corresponding arguments. $ActType$ indicates the type of action $a_i$ corresponding to $w_i$, which can be one of \emph{essential}, \emph{optional} and \emph{exclusive}. The type $essential$ suggests the corresponding action $a_i$ to be extracted, $optional$ suggests $a_i$ that can be ``optionally'' extracted, $exclusive$ suggests $a_i$ that is ``exclusive'' with other actions indicated by the set $\{ExActId\}$ (in other words, either $a_i$ or exactly one action in $\{ExActId\}$ can be extracted). $ExActId$ is the index of the action exclusive with $a_i$. We denote the size of $\{ExActId\}$ by $M$, i.e., $|\{ExActId\}|=M$. Note that ``$M=0$'' indicates the type $ActType$ of action $a_i$ is either \emph{essential} or \emph{optional}, and ``$M\neq 0$'' indicates $ActType$ is \emph{exclusive}.
$ArgId$ is the index of the word composing arguments of $a_i$, and $ExArgId$ is the index of words exclusive with $ArgId$.

\begin{figure}[!ht]
\begin{center}
\includegraphics[width=0.45\textwidth]{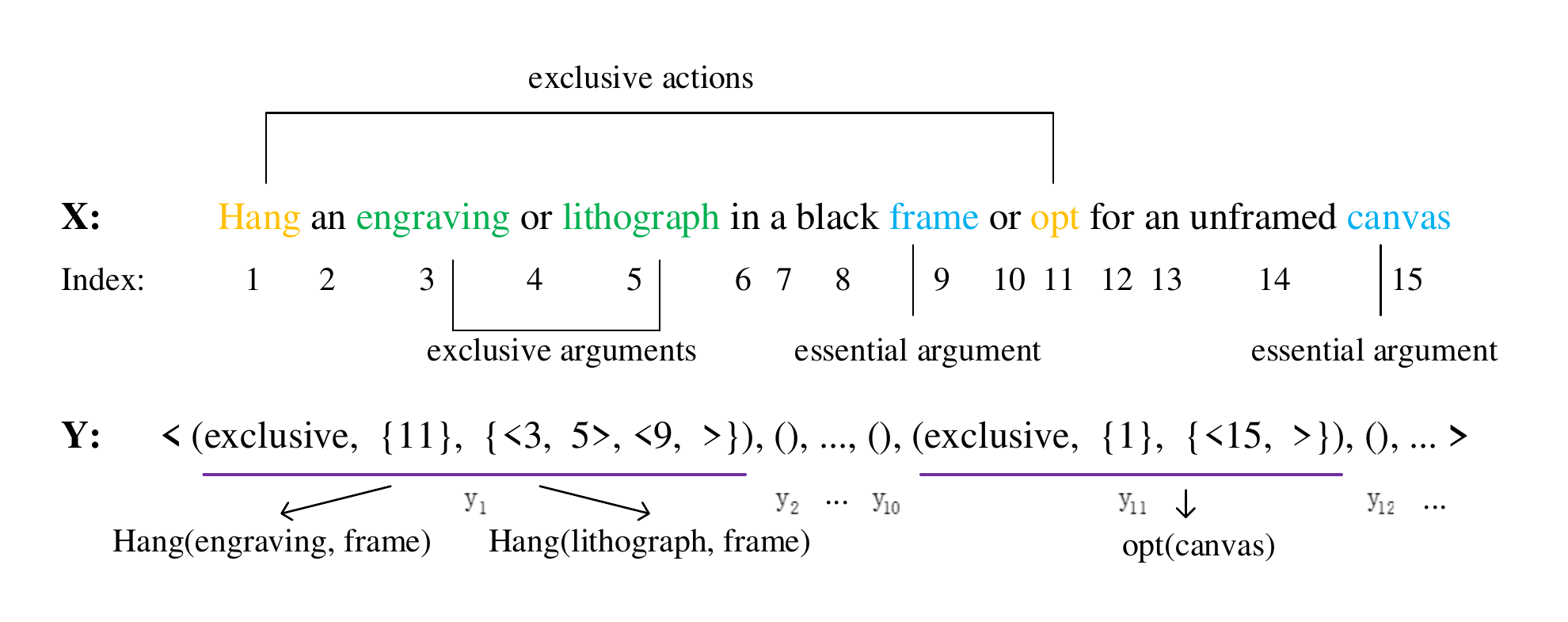}
\caption{Illustration of text X and its corresponding annotation Y}
\label{exampleXY}
\end{center}
\end{figure}

\emph{
For example, as shown in Figure \ref{exampleXY}, given a text denoted by $X$, its corresponding annotation is shown in the figure denoted by $Y$. In $y_1$, ``\{11\}'' indicates the action exclusive with $w_1$ (i.e., ``Hang'') is ``opt'' with index 11. ``$\{\langle 3,5\rangle,\langle 9,\rangle\}$'' indicates the corresponding arguments ``engraving'' and ``lithograph'' are exclusive, and the other argument ``frame'' with index 9 is essential since it is exclusive with an empty index, likewise for $y_{11}$. For $y_2,\ldots,y_{10}$ and $y_{12},\ldots,y_{15}$, they are empty since their corresponding words are not action names. From $Y$, we can generate three possible actions as shown at the bottom of Figure \ref{exampleXY}.
}

As we can see from the training data, it is uneasy to build a supervised learning model to directly predict annotations for new texts $X$, since annotations $y_i$ is complex and the size $|y_i|$ varies with respect to different $w_i$ (different action names have different arguments with different lengths). We seek to build a \emph{unified} framework to predict \emph{simple} ``labels'' (corresponding to ``actions'' in reinforcement learning) for extracting action names and their arguments. We exploit the framework to learn two models to predict action names and arguments, respectively. Specifically, given a new text $X$, we would like to predict a sequence of operations $O=\langle o_1,o_2,\ldots,o_N\rangle$ (instead of annotations in $\Phi$) on $X$, where $o_i$ is an $operation$ that $selects$ or $eliminates$ word $w_i$ in $X$. In other words, when predicting action names (or arguments), $o_i=Select$ indicates $w_i$ is extracted as an action name (or argument), while $o_i=Eliminate$ indicates $w_i$ is not extracted as an action name (or argument). 

In summary, our action sequence extraction problem can be defined by: given a set of training data $\Phi$, we aim to learn two models (with the same framework) to predict action names and arguments for new texts $X$, respectively. The two models are 
\begin{equation}\label{m1}
\mathcal{F}^1_{\Phi}(O|X; \theta_1)
\end{equation}
and 
\begin{equation}\label{m2}
\mathcal{F}^2_{\Phi}(O|X,a; \theta_2),
\end{equation}
where $\theta_1$ and $\theta_2$ are parameters to be learnt for predicting action names and arguments, respectively. $a$ is an action name extracted based on $\mathcal{F}^1_{\Phi}$. We train $\mathcal{F}^2_{\Phi}$ for extracting arguments based on ground-truth action names. When testing, we extract arguments based on the action names extracted by $\mathcal{F}^1_{\Phi}$. We will present the details of building these two models in the following sections.

\section{Our {\ours} Approach}
In this section we present the details of our {\ours} approach. As mentioned in the introduction section, our action sequence extraction problem can be viewed as a reinforcement learning problem. We thus first describe how to build \emph{states} and \emph{operations} given text $X$, and then present deep Q-networks to build the Q-functions. Finally we present the training procedure and give an overview of our {\ours} approach. Note that we will use the term \emph{operation} to represent the meaning of ``action'' in reinforcement learning since the term ``action'' has been used to represent an action name with arguments in this work.

\subsection{Generating State Representations}
\begin{figure}[!ht]
\begin{center}
\includegraphics[width=0.38\textwidth]{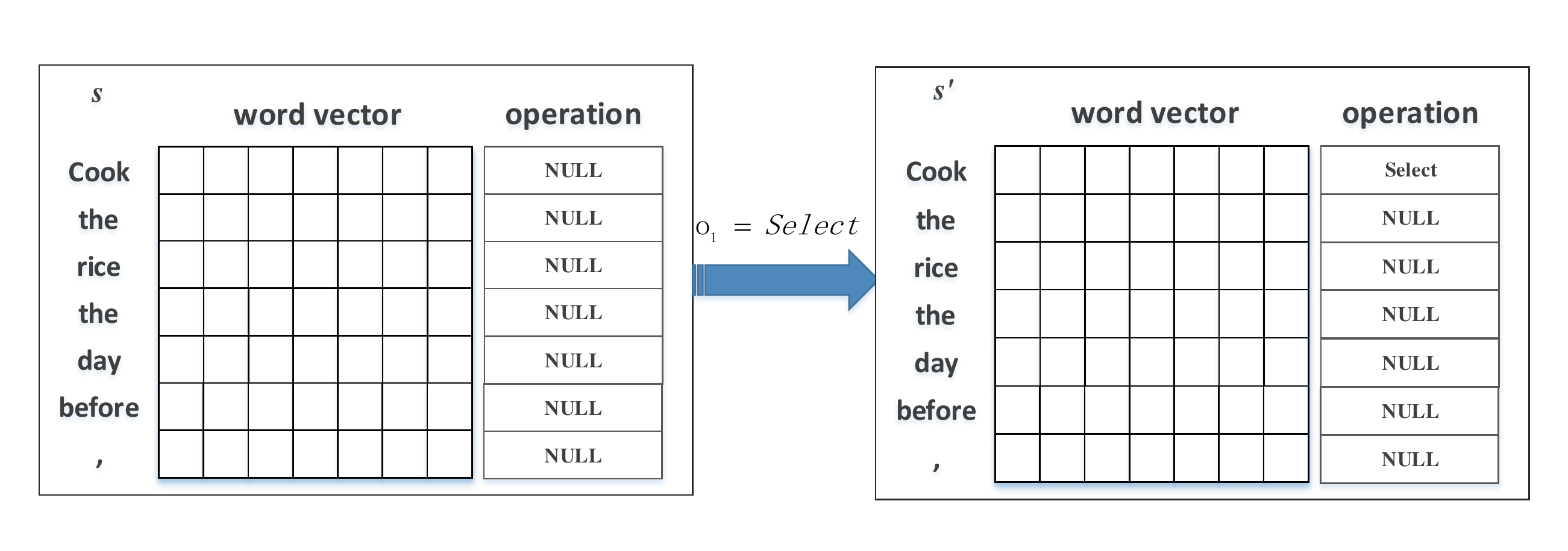}
\caption{Illustration of states and operations}
\label{text_matrix}
\end{center}
\end{figure}

In this subsection we address how to generate state representations from texts.
As defined in the problem definition section, the space of operations is $\{Select, Eliminate\}$. We view texts associated with operations as ``states''. Specifically, we represent a text $X$ by a sequence of vectors $\langle \w_1,\w_2,\ldots,\w_N\rangle$, where $\w_i \in \mathcal{R}^{K_1}$ is a $K_1$-dimension real-valued vector \cite{word2vec}, representing the $i$th word in $X$. Words of texts stay the same when we perform operations,  so we embed operations in state representations to generate state transitions. We extend the set of operations to $\{NULL,Select,Eliminate\}$ where ``NULL'' indicates a word has not been processed. We represent  the operation sequence $O$ corresponding to $X$ by a sequence of vectors $\langle \mathbf{o}_1,\mathbf{o}_2, \ldots, \mathbf{o}_N\rangle$, where $\mathbf{o}_i \in \mathcal{R}^{K_2}$ is a $K_2$-dimension real-valued vector. In order to balance the dimension of $\mathbf{o}_i$ and $\w_i$, we generate each $\mathbf{o}_i$ by a \emph{repeat-representation} $[\cdot]_{K_2}$, i.e., if $K_2=1$, $\mathbf{o}_i\in\{[0],[1],[2]\}$, and if $K_2=3$, $\mathbf{o}_i\in\{[0,0,0],[1,1,1],[2,2,2]\}$, where $\{0,1,2\}$ corresponds to $\{NULL,Select,Eliminate\}$, respectively. 
We define a \emph{state} $s$ as a tuple $\langle \mathbf{X},\mathbf{O}\rangle$, where $\mathbf{X}$ is a matrix in $\mathcal{R}^{K_1\times N}$, $\mathbf{O}$ is a matrix in $\mathcal{R}^{K_2\times N}$. The $i$th row of $s$ is denoted by $[\w_i,\mathbf{o}_i]$. The space of states is denoted by $\mathcal{S}$. A state $s$ is changed into a new state $s'$ after performing an operation $\mathbf{o}'_i$ on $s$, such that $s'=\langle \X,\mathbf{O}'\rangle$, where $\mathbf{O}'=\langle \mathbf{o}_1,\ldots,\mathbf{o}_{i-1},\mathbf{o}'_i,\mathbf{o}_{i+1},\ldots,\mathbf{o}_N\rangle$.
\emph{For example, consider a text ``Cook the rice the day before..." and a state $s$ corresponding to it is shown in the left part of Figure \ref{text_matrix}. After performing an operation $\mathbf{o}_1=Select$ on $s$, a new state $s'$ (the right part) will be generated.} In this way,  we can learn $\theta_1$ in $\mathcal{F}^1_{\Phi}$ (Equation (\ref{m1})) based on $s$ with deep Q-networks as introduced in the next subsection.

After $\mathcal{F}^1_{\Phi}$ is learnt, we can use it to predict action names, and then exploit the predicted action names to extract action arguments by training  $\mathcal{F}^2_{\Phi}$ (Equation (\ref{m2})). To do this, we would like to encode the predicted action names in states to generate a new state representation $\hat{s}$ for learning $\theta_2$ in $\mathcal{F}^2_{\Phi}$. We denote by $w_a$ the word corresponding to the action name. We build $\hat{s}$ by appending the distance between $w_a$ and $w_j$ based on their indices, such that $\hat{s}=\langle \mathbf{X},\mathbf{D},\mathbf{O}\rangle$, where $\mathbf{D}=\langle \mathbf{d}_1,\mathbf{d}_2,\ldots, \mathbf{d}_N\rangle$, where $\mathbf{d}_j=[d_j]_{K_3}$ and $d_j=|a-j|$. Note that $\mathbf{d}_j$ is a $K_3$-dimension real-valued vector using \emph{repeat-representation} $[\cdot]_{K_3}$. In this way we can learn $\mathcal{F}^2_{\Phi}$ based on $\hat{s}$ with the same deep Q-networks. 
Note that in our experiments, we found that the results were the best when we set $K_1=K_2=K_3$, suggesting the impact of word vectors, distance vectors and operation vectors was generally identical.

\subsection{Deep Q-networks for Operation Execution}
Given the formulation of states and operations, we aim to extract action sequences from texts. We construct sequences by repeatedly choosing operations given current states, and applying operations on current states to achieve new states. 

In Q-Learning, this process can be described by a Q-function and updating the Q-function iteratively according to Bellman equation. In our action sequence extraction problem, actions are composed of action names and action arguments. We need to first extract action names from texts and use the extracted action names to further extract action arguments. Specifically, we define two Q-functions $Q(s,o)$ and $Q(\hat{s},o)$, where $\hat{s}$ contains the information of extracted action names, as defined in the last subsection. The update procedure based on Bellman equation and deep Q-networks can be defined by:

\begin{equation}
Q_{i+1}(s,o;\theta_1) = E\left\lbrace r+\gamma\max_{o'}Q_i(s',o';\theta_1)|s,o\right\rbrace
\label{Bellman}
\end{equation}
\begin{equation}
Q_{i+1}(\hat{s},o;\theta_2) = E\left\lbrace r+\gamma\max_{o'}Q_i(\hat{s}',o';\theta_2)|\hat{s},o\right\rbrace
\label{Bellman}
\end{equation}
where $Q_{i+1}(s,o;\theta_1)$ and $Q_{i+1}(\hat{s},o;\theta_2)$ correspond to the deep Q-networks \cite{Mnih2015Human} for extracting action names and arguments, respectively. As $i \rightarrow \infty$, $Q_i \rightarrow Q^*$. In this way, we can define $\mathcal{F}^1_{\Phi}=Q^*(s,o;\theta_1)$ and $\mathcal{F}^2_{\Phi}=Q^*(\hat{s},o;\theta_2)$ in Equations (\ref{m1}) and (\ref{m2}), and then use $\mathcal{F}^1_{\Phi}$ and $\mathcal{F}^2_{\Phi}$ to extract action names and arguments, respectively.

Since Convolutional Neural Networks (CNNs) are widely applied in natural language processing \cite{DBLP:conf/emnlp/Kim14,DBLP:journals/corr/ZhangW15b,ijcai2017-406}, we build CNN models to learn Q-functions $Q(s,o,\theta_1)$ and $Q(\hat{s},o,\theta_2)$. We adopt the CNN Architecture of \cite{DBLP:journals/corr/ZhangW15b}. To build the kernels of our CNN models, we test from uni-gram context to ten-gram context and observe that five-word context works well in our task. We thus design four types of kernels, which correspond to bigram, trigram, four-gram and five-gram, respectively.

\subsection{Computing Rewards}
In this subsection we compute the reward $r$ based on state $s$ and operation $o$. Specifically, $r$ is composed of two parts, i.e., \textbf{basic reward} and \textbf{additional reward}. For the basic reward at time step $\tau$, denoted by $r_{b,\tau}$, if a word is not an \emph{item} (we use \emph{item} to represent action name or action argument when it is not confused), $r_{b,\tau}$ is $+50$ when the operation is correct and $-50$ otherwise. If a word is an \emph{essential item}, $r_{b,\tau}=+100$ when the operation is correct and $r_{b,\tau}=-100$ when it is incorrect. If the word is an \emph{optional item}, $r_{b,\tau}=+100$ when the operation is correct and $r_{b,\tau}=0$ when it is incorrect. If a word is an \emph{exclusive item}, $r_{b,\tau}=+150$ when the operation is correct and $r_{b,\tau}=-150$ when it is incorrect. We denote that an operation is correct when it selects essential items, selects optional items, selects only one item of exclusive items or eliminates words that are not items.

Note that action names are key verbs of a text and action arguments
are some nominal words, so the percentage of these words in a text is
closely related to action sequence extraction process.  We thus calculate
the percentage, namely an \emph{item rate}, denoted by
$\delta=\frac{\#Item}{\#Word}$, where $\#Item$ indicates the amount of
action names or action arguments in all the annotated texts and
$\#Word$ indicates the total number of words of these texts. We define a \emph{real-time item rate} as $\delta_\tau$ to denote the percentage of words that have been selected as action names or action arguments in a text \emph{after $\tau$ training steps}, and $\delta_0=0$. On one hand, when $\delta_{\tau-1} \leq \delta$, a positive additional reward is added to $r_{b,\tau}$ if $r_{b,\tau} \geq 0$ (i.e., the operation is correct), otherwise a negative additional reward is added to $r_{b,\tau}$. On the other hand, when $\delta_\tau > \delta$, which means that words selected as action names or action arguments are out of the expected number and it is more likely to be incorrect if subsequent words are selected, then a negative additional reward should be added to the basic reward. In this way, the reward $r_\tau$ at time step $\tau$ can be obtained by Equation (\ref{complete reward}),
\begin{equation}
r_\tau=\begin{cases}
r_{b,\tau}+\sgn{r_{b,\tau}} \cdot c\delta_{\tau-1} & \text{$\delta_{\tau-1} \leq \delta$},\\
r_{b,\tau}-c\delta_{\tau-1} & \text{$\delta_{\tau-1} > \delta$}.
\end{cases}
\label{complete reward}
\end{equation}
where $c$ is a positive constant and $0 \leq \delta_{\tau-1} < 1$.  \ignore{We empirically use $c\delta_\tau$ to represent the additional reward, which can be replaced by others. It is, however, effective based on our experiments. }

\subsection{Training Our Model}
To learn the parameters $\theta_1$ and $\theta_2$ of our two DQNs, we
store transitions $\langle s,o,r,s'\rangle$ and  $\langle
\hat{s},o,r,\hat{s}'\rangle$ in replay memories $\Omega$ and
$\hat{\Omega}$, respectively, and exploit a mini-batch sampling
strategy. As indicated in \cite{DBLP:conf/emnlp/NarasimhanKB15},
transitions that provide positive rewards can be used more often to
learn optimal Q-values faster. We thus develop a \emph{positive-rate
  based experience replay} instead of randomly sampling transitions
from $\Omega$ (or $\hat{\Omega}$), where \emph{positive-rate}
indicates the percentage of transitions with positive rewards. To do
this, we set a positive rate $\rho (0 < \rho < 1 )$ and require the
proportion of positive samples in each mini-batch be $\rho$.

We present the learning procedure of our {\ours} approach in Algorithm \ref{our_algorithm}, for building $\mathcal{F}^1_{\Phi}$. We can simply replace $s_1$, $\Omega$ and $\theta_1$ with $\hat{s}_1$, $\hat{\Omega}$ and $\theta_2$ for building $\mathcal{F}^2_{\Phi}$. In Step 4 of Algorithm \ref{our_algorithm}, we generate the initial state $s_1$ ($\hat{s}_1$ for learning $\mathcal{F}^2_{\Phi}$) for each training data $\Phi=\{\langle X,Y\rangle\}$ by setting all operations $o_i$ in $s_1$ to be $NULL$. We perform $N$ steps to execute one of the operations $\{Select, Eliminate\}$ in Steps 6, 7 and 8. From Steps 10 and 11, we do a \emph{positive-rate based experience replay} according to positive rate $\rho$. From Steps 12 and 13, we update parameters $\theta_1$ using gradient descent on the loss function $\mathcal{L}(\theta_1)=(y_j - Q(s_j,o_j; \theta_1))^2$ as shown in Step 13.

With Algorithm \ref{our_algorithm}, we are able to build the Q-function $Q(s,o;\theta_1)$  and execute operations $\{Select, Eliminate\}$ to a new text by iteratively maximizing the Q-function. Once we obtain operation sequences, we can generate action names and use the action names to build $Q(\hat{s},o;\theta_2)$ with $\hat{\Omega}$ and the same framework of Algorithm \ref{our_algorithm}. We then exploit the built $Q(\hat{s},o;\theta_2)$ to extract action arguments. As a result, we can extract action sequences from texts using both of the built $Q(s,o;\theta_1)$ and $Q(\hat{s},o;\theta_2)$.

\begin{algorithm}[!ht]
\caption{Our {\ours} algorithm}
\label{our_algorithm}
\textbf{Input:} a training set $\Phi$, positive rate $\rho$, item rate $\delta$  \\
\textbf{Output:} the parameters $\theta_1$ 

\begin{algorithmic}[1]
\STATE Initialize $\Omega=\emptyset$, CNN with random values for $\theta_1$
\FOR{epoch = 1: $H$} 
     \FOR {each training data $\langle X,Y\rangle\in\Phi$}
	\STATE Generate the initial state $s_1$ based on $X$
	\FOR{$\tau$ = 1: $N$}
		\STATE Perform an operation $o_\tau$ with probability $\epsilon$  
		\STATE Otherwise select $o_\tau = \max\limits_{o} Q(s_\tau, o; \theta_1)$
		\STATE Perform $o_\tau$ on $s_\tau$ to generate $s_{\tau+1}$
		\STATE Calculate $r_\tau$ based on $s_{\tau+1}$, $o_\tau$, $Y$ and $\delta$
		\STATE Store transition $(s_\tau, o_\tau, r_\tau, s_{\tau+1})$ in $\Omega$
		\STATE Sample $(s_j,o_j,r_j,s_{j+1})$ from $\Omega$ based on $\rho$
		\STATE Set \\$y_j=\begin{cases} r_j \qquad\qquad\qquad \text{for terminal $s_{j+1}$} \\ r_j + \gamma\max\limits_{o'}Q(s_{j+1},o';\theta_1) \ \text{otherwise} \end{cases}$
		\STATE Update $\theta_1$ based on loss function $\mathcal{L}(\theta_1)$
	\ENDFOR
      \ENDFOR
\ENDFOR
\RETURN The parameters $\theta_1$ 
\end{algorithmic}
\end{algorithm}

\section{Experiments}
\subsection{Datasets and Evaluation Metric}
We conducted experiments on three datasets, i.e., ``Microsoft Windows
Help and Support'' (WHS) documents \cite{DBLP:conf/acl/BranavanCZB09},
and two datasets collected from ``WikiHow Home and
Garden''\footnote{https://www.wikihow.com/Category:Home-and-Garden}
(WHG) and ``CookingTutorial''\footnote{http://cookingtutorials.com/}
(CT). Details are presented in Table \ref{Statistics of
  Datasets}. Supervised learning models require that training data are
one-to-one pairs (i.e. each word has a unique label), so we generate
input-texts-to-output-labels based on annotation $Y$ (as defined in
Section \ref{def}). In our task, a single text with $n$ optional items
or $n$ exclusive pairs can  generate more than $2^n$ potential label sequences (i.e. each item of them can be extracted or not be extracted). Especially, we observe that $n$ is larger than 30 in some texts of our datasets, which means more than 1 billion sequences will be generated. We thus restrict $n\leq 8$ (no more than $2^{8}$ label sequences) to generate reasonable number of sequences. 
\begin{table}[!htbp]
\caption{Datasets used in our experiments }
\label{Statistics of Datasets}
\centering
\begin{tabular}{lccc}
\hline
                        & \textbf{WHS} & \textbf{CT} & \textbf{WHG} \\
\hline
Labeled texts          & 154          & 116         & 150      \\
Input-output pairs   &1.5K        & 134K  & 34M  \\
Action name rate (\%)      & 19.47        & 10.37       & 7.61 \\
Action argument rate (\%)      & 15.45        & 7.44        & 6.30  \\
Unlabeled texts        & 0            & 0           & 80  \\
\hline
\end{tabular}
\end{table}

For evaluation, we first feed test texts to each model to output sequences of labels or operations. We then extract action sequences based on these labels or operations. After that, we compare these action sequences to their corresponding annotations and calculate $\#TotalTruth$ (total ground truth items), $\#TotalTagged$ (total extracted items), $\#TotalRight$ (total correctly extracted items). Finally we compute metrics: $precision=\frac{\#TotalRight}{\#TotalTagged}$, $recall=\frac{\#TotalRight}{\#TotalTruth}$, and $F1=\frac{2\times precision \times recall}{precision  + recall}$. We randomly split each dataset into 10 folds, calculated an average of performance over 10 runs via 10-fold cross validation, and used the F1 metric to validate the performance in our experiments.

\subsection{Experimental Results}
We compare {\ours} to four baselines, as shown below:
\begin{itemize}
\item{STFC:} Stanford CoreNLP, an off-the-shelf tool,  denoted by STFC, extracts action sequences by viewing root verbs as action names and objects as action arguments \cite{DBLP:conf/aips/LindsayRFHPG17}.

\item{BLCC:}  Bi-directional LSTM-CNNs-CRF model \cite{DBLP:conf/acl/MaH16,DBLP:conf/emnlp/ReimersG17} is a state-of-the-art sequence labeling approach. We fine-tuned parameters of the approach, including character embedding, embedding size, dropout rate, etc., and denoted the resulting approach by BLCC.

\item{EAD:} The Encoder-Aligner-Decoder approach maps instructions to action sequences proposed by \cite{Mei2016Listen}, denoted by EAD. 

\item{CMLP:} We consider a Combined Multi-layer Perceptron (CMLP), which consists of $N$ MLP classifiers. $N=500$ for action names extraction and $N=100$ for action arguments extraction. Each MLP classifier focuses on  not only a single word but also the k-gram context. 
\end{itemize}

When comparing with baselines, we adopt the settings used by \cite{DBLP:journals/corr/ZhangW15b} to build our CNN networks. We set the input dimension to be $(500 \times 100)$ for action names and $(100 \times 150)$ for action arguments, the number of feature-maps to be $32$. We used $0.25$ dropout on the concatenated max pooling outputs and exploited a $256$ dimensional fully-connected layer before the final two dimensional outputs. We set the replay memory $\Omega=100000$, discount factor $\gamma=0.9$. We varied $\rho$ from $0.05$ to $0.95$ with the interval of $0.05$ and found the best value is $0.80$ (that is why we set $\rho=0.80$ in the experiment). We set $\delta=0.10$ for action names, $\delta=0.07$ for arguments according to Table \ref{Statistics of Datasets}, the constant $c =50$, learning rate of adam to be 0.001, probability $\epsilon$ for $\epsilon$-greedy decreasing from 1 to 0.1 over 1000 training steps. 

\subsubsection{Comparison with Baselines} 
\begin{table}[!ht]
\caption{F1 scores of different methods in extracting all types of action names and all types of action arguments}
\label{best reaults}
\begin{small}
\begin{tabular}{c|ccc|ccc}
\hline
 & \multicolumn{3}{c|}{\textbf{Action Names}} & \multicolumn{3}{c}{\textbf{Action Arguments}} \\
Method & \textbf{WHS} & \textbf{CT} & \textbf{WHG} & \textbf{WHS} & \textbf{CT} & \textbf{WHG} \\
\hline
EAD-2  & 86.25 & 64.74 & 53.49 & 57.71 & 51.77 & 37.70 \\
EAD-8  & 85.32 & 61.66 & 48.67 & 57.71 & 51.77 & 37.70 \\
CMLP-2 & 83.15 & 83.00 & 67.36 & 47.29 & 34.14 & 32.54 \\
CMLP-8 & 80.14 & 73.10 & 53.50 & 47.29 & 34.14 & 32.54 \\
BLCC-2 & 90.16 & 80.50 & 69.46 & 93.30 & \textbf{76.33} & 70.32 \\
BLCC-8 & 89.95 & 72.87 & 59.63 & 93.30 & \textbf{76.33} & 70.32 \\
STFC & 62.66 & 67.39 & 62.75 & 38.79 & 43.31 & 42.75 \\
{\ours} & \textbf{93.46} & \textbf{84.18} & \textbf{75.40} & 
          \textbf{95.07} & 74.80 & \textbf{75.02} \\
\hline
\end{tabular}
\end{small}
\end{table}
We set the restriction $n=2$ and $n=8$ for EAD, CMLP and BLCC which need one-to-one sequence pairs, and no restriction for STFC and {\ours}. In all of our datasets, the arguments of an action are either all essential arguments or one exclusive argument pair together with all other essential arguments, which means at most $2^1$ sequences can be generated. Therefore, the results of action arguments extraction are identical when $n=2$ and $n=8$. The experimental results are shown in Table \ref{best reaults}. From Table \ref{best reaults}, we can see that our {\ours} approach performs the best on extracting both action names and action arguments in most datasets, except for CT dataset. We observe that the number of arguments in most texts of the CT dataset is very small, such that BLCC performs well on extracting arguments in the CT dataset.
On the other hand, we can also observe that BLCC, EAD and CMLP get
worse performance when relaxing the restriction on  $n$ ($n=2$ and $n=8$). The reason is that when given a single text with many possible output sequences, these models learn common parts (essential items) of outputs, neglecting the different parts (optional or exclusive items). We can also see that both sequence labeling method and encoder-decoder structure do not work well, which exhibits that, in this task, our reinforcement learning framework can indeed perform better than traditional methods.

\begin{table}[!ht]
\caption{F1 scores of different methods in extracting exclusive action names and exclusive action arguments}
\label{exclusive study}
\begin{small}
\begin{tabular}{c|ccc|ccc}
\hline
 & \multicolumn{3}{c|}{\textbf{Action Names}} & \multicolumn{3}{c}{\textbf{Action Arguments}} \\
Method & \textbf{WHS} & \textbf{CT} & \textbf{WHG} & \textbf{WHS} & \textbf{CT} & \textbf{WHG} \\
\hline
EAD-2  & 26.60 & 21.76 & 22.75 & 40.78 & 47.91 & 39.81 \\
EAD-8  & 22.12 & 17.01 & 23.12 & 40.78 & 47.91 & 39.81 \\
CMLP-2 & 31.54 & 54.75 & 51.29 & 35.52 & 25.07 & 29.78 \\
CMLP-8 & 26.90 & 51.80 & 41.03 & 35.52 & 25.07 & 29.78 \\
BLCC-2 & 16.35 & 38.27 & 54.34 & 12.50 & 13.45 & 18.57 \\
BLCC-8 & 19.55 & 35.01 & 41.27 & 12.50 & 13.45 & 18.57 \\
STFC   & 46.40 & 50.28 & 44.32 & 50.00 & 46.40 & 50.32 \\
{\ours} & \textbf{56.19} & \textbf{66.37} & \textbf{68.29} & 
          \textbf{66.67} & \textbf{54.24} & \textbf{55.67} \\
\hline
\end{tabular}
\end{small}
\end{table}

\begin{figure}[!ht]
\centering
\includegraphics[width=0.2\textwidth]{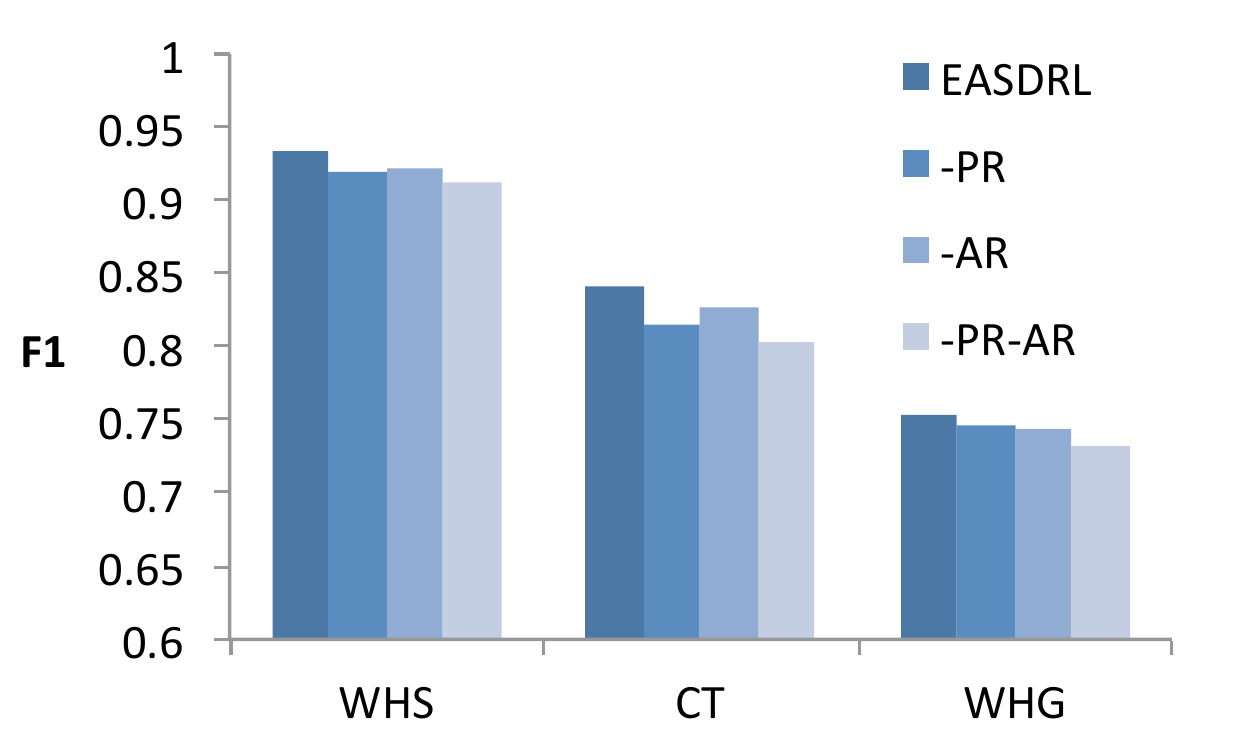}
\includegraphics[width=0.2\textwidth]{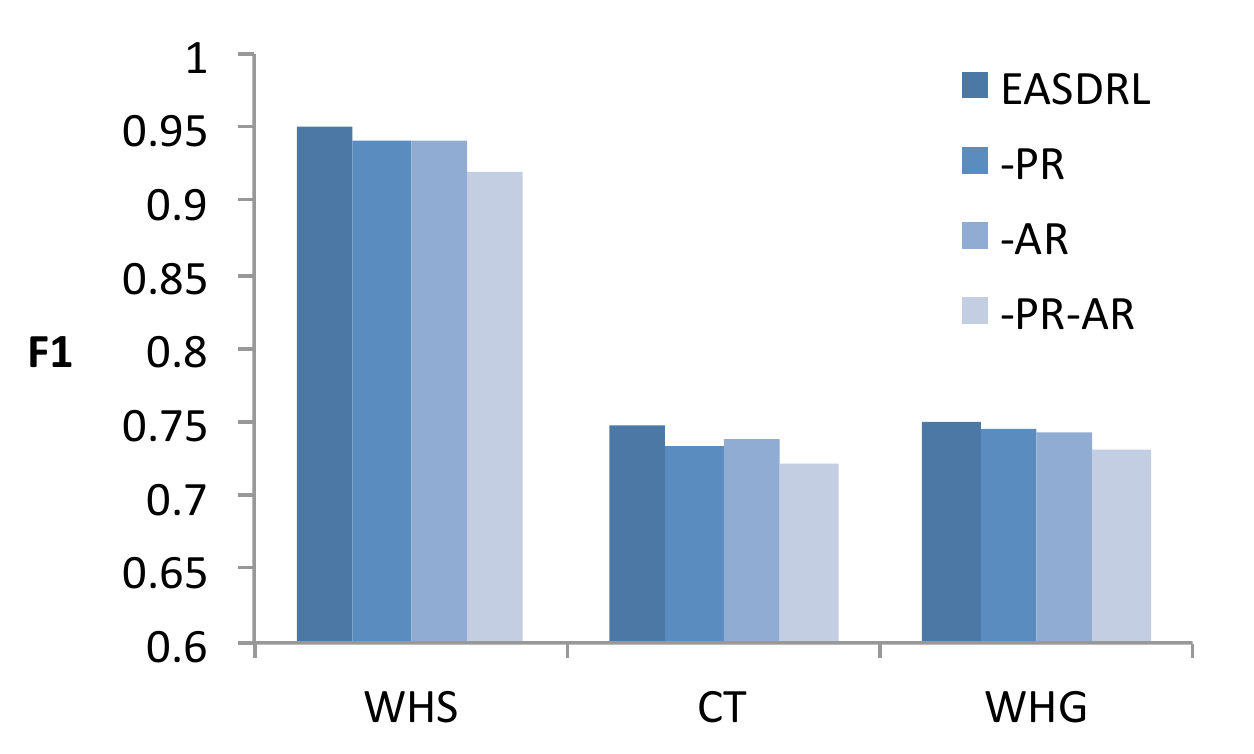}
\caption{Results of {\ours} ablation studies}
\label{ablation study}
\end{figure}

In order to test and verify whether or not our {\ours} method can deal with complex action types well, we compare with baselines in extracting \emph{exclusive action names} and \emph{exclusive action arguments}. Results are shown in Table \ref{exclusive study}. In this part, our {\ours} model outperforms all baselines and leads more than $5\%$ absolutely, which demonstrates the effectiveness of our {\ours} model in this task.

We would like to evaluate the impact of additional reward and positive-rate based experience replay. We test our {\ours} model by removing  \emph{positive-rate based experience replay} (denoted by ``-PR'') or \emph{additional reward} (denoted by ``-AR''). Results are shown in Figure \ref{ablation study}. We observe that removing either \emph{positive-rate based experience replay} or \emph{additional reward} degrades the performance of our model. 

\subsubsection{Online Training Results}
\begin{figure}[!ht]
\centering
\includegraphics[width=0.48\textwidth]{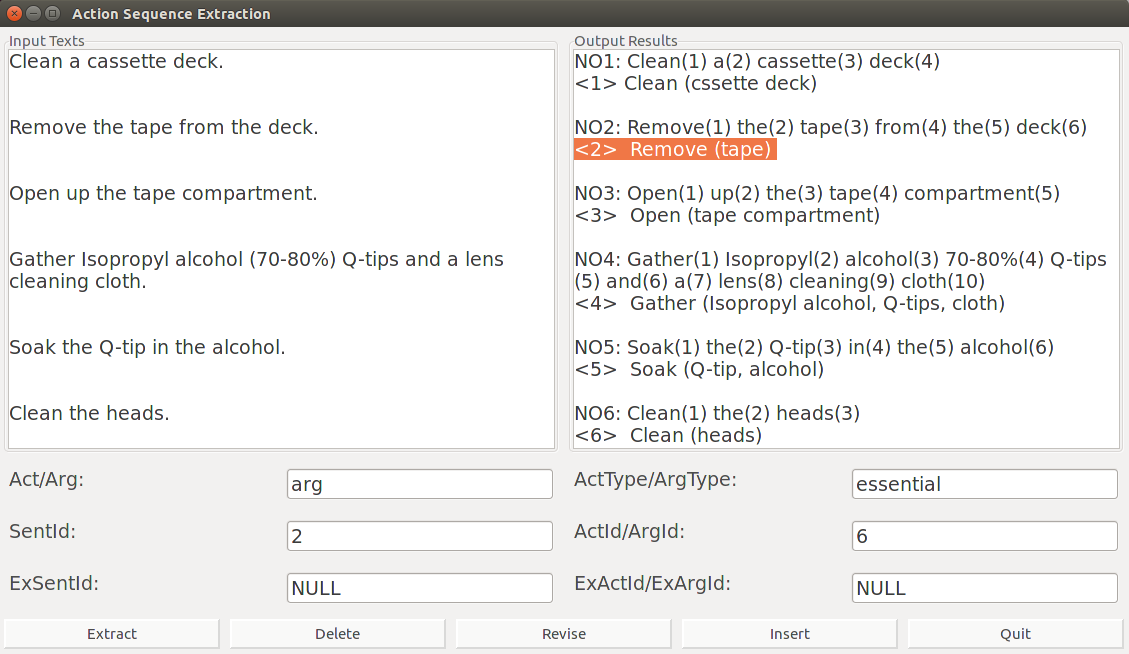}
\caption{A snapshot of our human-agent interacting environment}
\label{gui}
\end{figure}

To further test the robustness and self-learning ability of our approach, we design a human-agent interaction environment to collect the feedback from humans. The environment takes a text as input (as shown in the upper left part of Figure \ref{gui}) and present the results of our {\ours} approach in the upper right part of Figure \ref{gui}. Humans adjust the output results by inputting values in the ``function panel'' (as shown in the middle row) and pressing the buttons (in the bottom). After that, the environment updates the deep Q-networks of our {\ours} approach based on humans' adjustment (or feedback) and output new results in the upper right part. Note that the parts indicated by $\langle 1\rangle,\langle 2\rangle,\ldots, \langle 6\rangle$ in the upper right part comprise the extracted action sequence. For example, the action ``Remove(tape)'', which is indicated in the upper right part with orange color, should be ``Remove(tape, deck)''. The user can delete, revise or insert words (corresponding to the buttons with labels ``Delete'', ``Revise'' and ``Insert'', respectively) by input ``values'' in the middle row, where ``Act/Arg'' is used to decide whether the inputed words belong to action names or action arguments, ``ActType/ArgType'' is used to decide whether the inputed words are essential, optional or exclusive, ``SentId'' and ``ActId/ArgId'' are used to input the sentence indices and word indices of inputed words, ``ExSentId'' and ``ExActId/ExArgId'' are used to input the indices of exclusive action names or arguments. After that, the modified text with its annotations will be used to update our model.

\begin{figure}[!ht]
\centering
\includegraphics[width=0.2\textwidth]{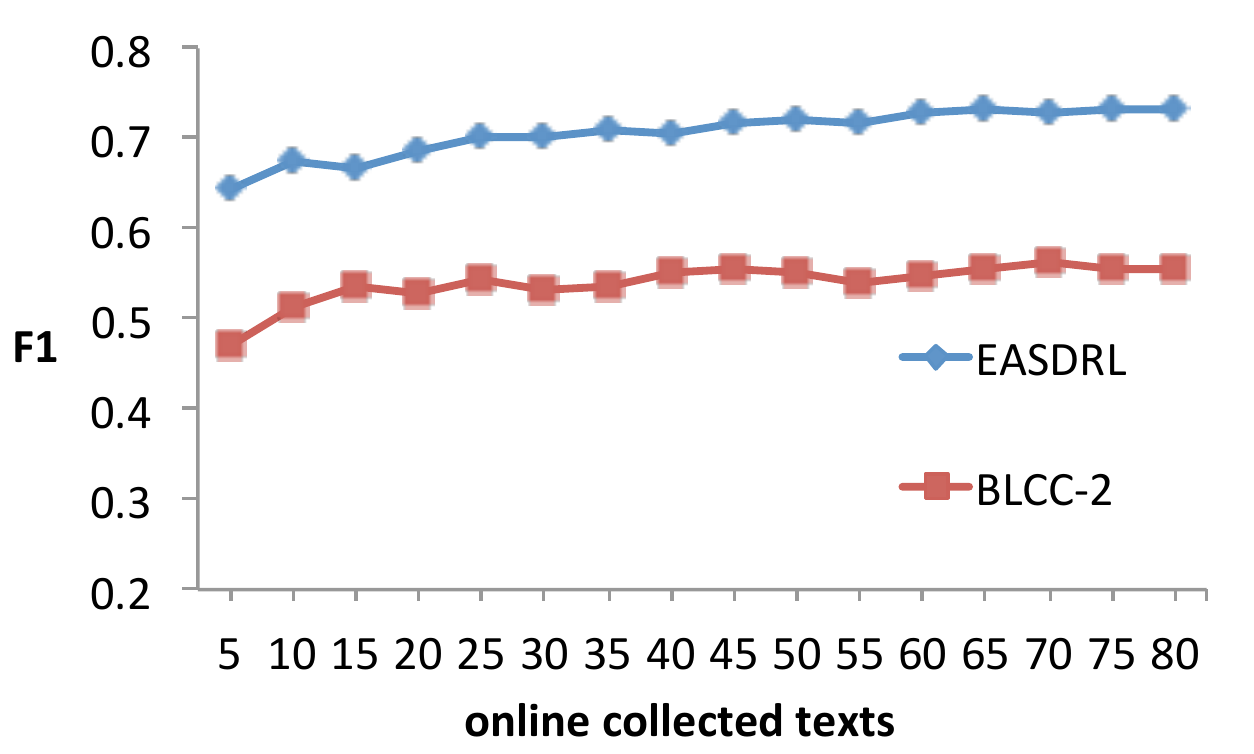}
\includegraphics[width=0.2\textwidth]{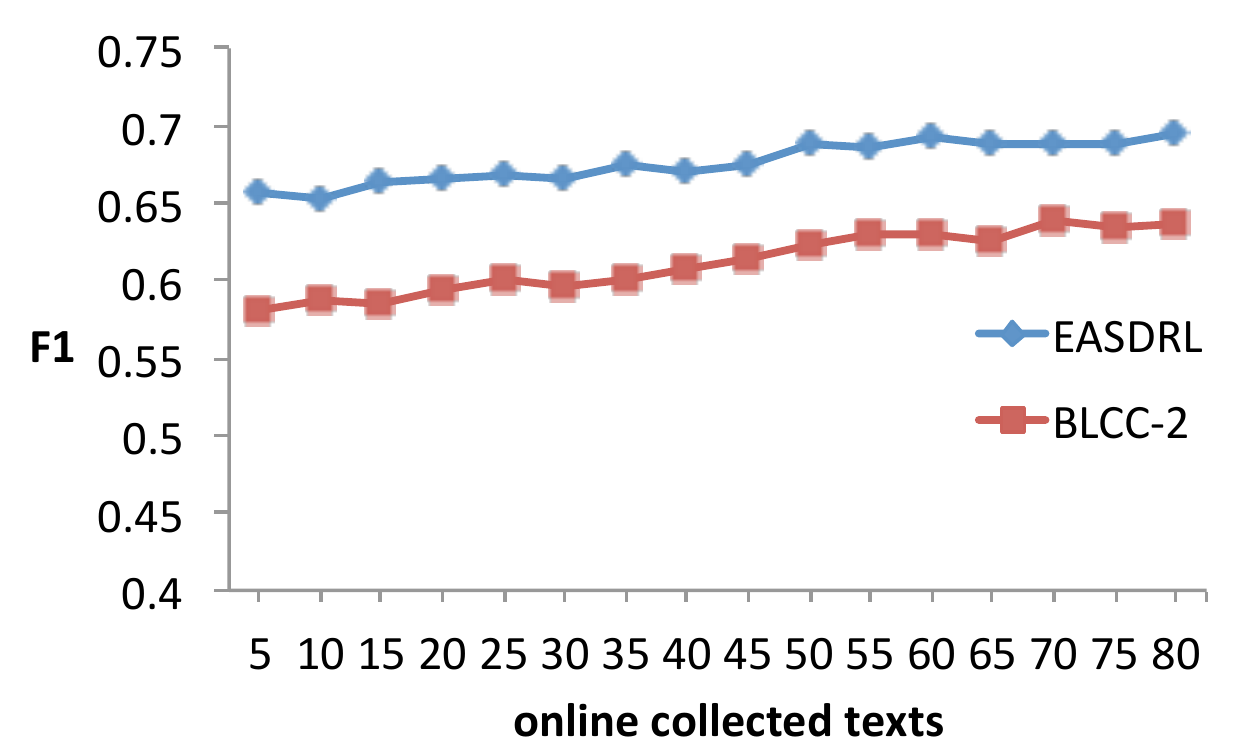}
\caption{Online test results of WHG dataset}
\label{online test}
\end{figure}

Before online training, we pre-train an initial model of {\ours} by combining all labeled texts of WHS, CT and WHG, with $30$ labeled texts of WHG for testing. The accuracy of this initial model is low since it is domain-independent. We then use the unlabeled texts in WHG (i.e., 80 texts as indicated in the last row in Table \ref{Statistics of Datasets}) for online training. We ``invited'' humans to provide feedbacks for these 80 texts (with an average of 5 texts for each human). When a human finishes the job assigned to him, we update our model (as well as the baseline model). We compare {\ours} to the best offline-trained baseline BLCC-2. Figure \ref{online test} shows the results of online training, where ``online collected texts'' indicates the number of texts on which humans provide feedbacks. We can see that {\ours} outperforms BLCC-2 significantly, which demonstrates the effectiveness of our reinforcement learning framework.

\section{Conclusion}
In this paper, we proposed a novel approach {\ours} to automatically
extract action sequences from texts based on deep reinforcement
learning. To the best of our knowledge, our {\ours} approach is the
first approach that explores deep reinforcement learning to extract
action sequences from texts. We empirically  demonstrated that our {\ours} model outperforms state-of-the-art baselines on three datasets. We showed that our {\ours} approach could better handle complex action types and arguments. We also exhibited the effectiveness of our {\ours} approach in an online learning environment. In the future, it would be interesting to explore the feasibility of learning more structured knowledge from texts such as state sequences or action models for supporting planning. 

\section*{Acknowledgements}
Zhuo thanks the support of the National Key Research and Development
Program of China (2016YFB0201900), National Natural Science Foundation
of China (U1611262), Guangdong Natural Science Funds for Distinguished
Young Scholar (2017A030306028), Pearl River Science and Technology New
Star of Guangzhou, and Guangdong Province Key Laboratory of Big Data
Analysis and Processing for the support of this research.
Kambhampati's research is supported in part by the AFOSR grant FA9550-18-1-0067, ONR grants N00014161-2892, N00014-13-1-0176, N00014- 13-1-0519, N00014-15-1-2027, and the NASA grant NNX17AD06G.

\bibliographystyle{named}
\bibliography{EASDRL}

\end{document}